\documentclass[letterpaper, 10 pt, conference]{ieeeconf}  

\IEEEoverridecommandlockouts                              

\usepackage{amsmath} 
\usepackage{amssymb}  
\usepackage{amsfonts}
\usepackage{graphicx}
\usepackage{url}
\usepackage[font={rm,md,up}]{subfig}
\usepackage{amssymb}
\usepackage{caption}
\usepackage{xcolor}
\usepackage{multirow}

\usepackage{changes}
\usepackage{float}
   
\title{\LARGE \bf
Generalizable Adversarial Examples Detection Based on Bi-model Decision Mismatch}

\author{Jo\~ao Monteiro$^{1,*}$, Isabela Albuquerque$^{1}$, Zahid Akhtar$^{2}$ and Tiago H. Falk$^{1}$ \\ 
$^{1}${\it{Institut National de la Recherche Scientifique, Montreal, Canada}} \\
$^{2}${\it{University of Memphis, Tennessee, USA}}\\
$^{*}${\tt\small joaomonteirof@gmail.com}}

\begin{document}

\maketitle

\begin{abstract}
Modern applications of artificial neural networks have yielded remarkable performance gains in a wide range of tasks. However, recent studies have discovered that such modelling strategy is vulnerable to \emph{Adversarial Examples}, i.e. examples with subtle perturbations often too small and imperceptible to humans, but that can easily fool neural networks. Defense techniques against adversarial examples have been proposed, but ensuring robust performance against varying or novel types of attacks remains an open problem. In this work, we focus on the detection setting, in which case attackers become identifiable while models remain vulnerable. Particularly, we employ the decision layer of independently trained models as features for posterior detection. The proposed framework does not require any prior knowledge of adversarial examples generation techniques, and can be directly employed along with unmodified off-the-shelf models. Experiments on the standard MNIST and CIFAR10 datasets deliver empirical evidence that such detection approach generalizes well across not only different adversarial examples generation methods but also quality degradation attacks. Non-linear binary classifiers trained on top of our proposed features can achieve a high detection rate ($>90\%$) in a set of white-box attacks and maintain such performance when tested against unseen attacks.
\end{abstract}

\section{Introduction}
\label{sec:introduction}

\begin{figure}[t]
  \centering
  \includegraphics[width=0.5\textwidth]{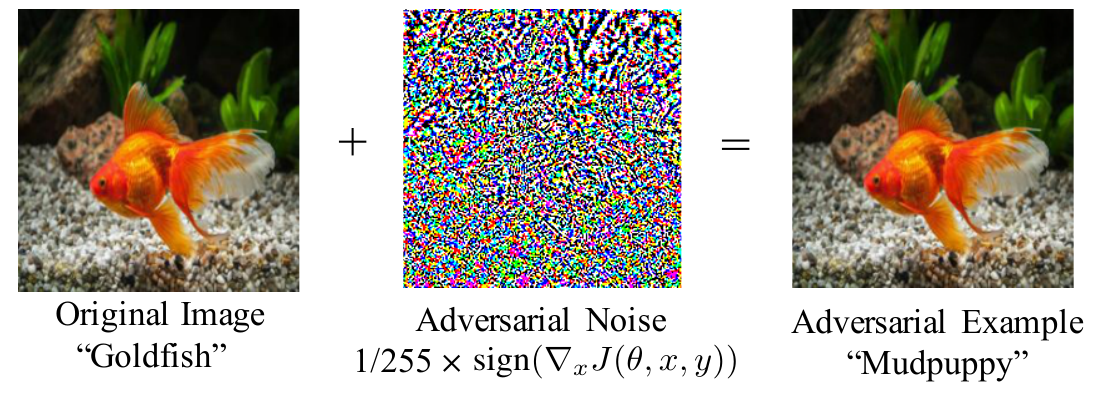}
  \caption{An adversarial sample generated with Fast Gradient Sign Method (FGSM) in \cite{Goodfellow2015}. VGG16 \cite{Simonyan2015} recognizes the clean original image sample correctly with high confidence. When a small perturbation is added to the image, the model predicts incorrect label with similar high confidence.}   
  \label{fig:AEexample}        
\end{figure}

Despite their major breakthroughs in solving challenging tasks, recent literature has shown that artificial neural networks \cite{goodfellow2016deep} can be vulnerable to deliberate perturbations, which, when added to input examples, can lead to incorrect predictions with high confidence \cite{Goodfellow2015}. Such perturbed examples are usually referred to as \emph{adversarial attacks}, i.e., carefully crafted variations of genuine samples intentionally modified so as to confuse or fool undefended models. For the particular case of neural networks, minimal perturbations are often required to yield effective adversarial attacks. As such, adversarial samples are often imperceptible to humans, but they result in model errors with high probability \cite{Moosavi2016}. For instance, an adversarial example as depicted in Fig. \ref{fig:AEexample} can be generated by adding some indiscernible perturbations into a given image. The resultant adversarial image is misclassified by the well-known convolutional classifier VGG16 \cite{Simonyan2015}, while a human being can still classify it correctly without spotting the deliberate added perturbations.

The vulnerability to adversarial attacks is severe since artificial neural networks are now being deployed in \emph{real-world} applications. Moreover, such attacks have been shown to be transferable \cite{Liu2017}. More specifically, attackers targeting a particular model can be utilized to compromise other models that have been designed for the same task, but with different model architectures and/or training techniques. As such, attackers could be designed targeting a known surrogate neural network and later applied to fool models practically deployed. This vulnerability can be particularly threatening in safety- and security-critical systems.

Following the description of adversarial attacks as first introduced in the work of Szegedy \emph{et al.} \cite{Szegedy2013}, several countermeasure techniques have been proposed since; these can be roughly categorized as \emph{defense} or \emph{detection} methods. Defense methods aim at improving the robustness of neural networks against attacks, i.e., make it either computationally more costly or require larger amounts of distortion imposed on the original example so as to yield an effective attack. Well known examples of defense strategies include adversarial training (i.e. including attacks in the training dataset) \cite{Goodfellow2015}, as well as training a new robust network to predict the outputs of a previously trained model, which is also known as the distillation method \cite{Papernot2016a}. Detection approaches, in turn, focus on being able to determine whether a given input is genuine or not, while leaving the model undefended \cite{Feinman2017,Grosse2017a}.

\begin{figure}
\centering
\includegraphics[width=0.9\columnwidth]{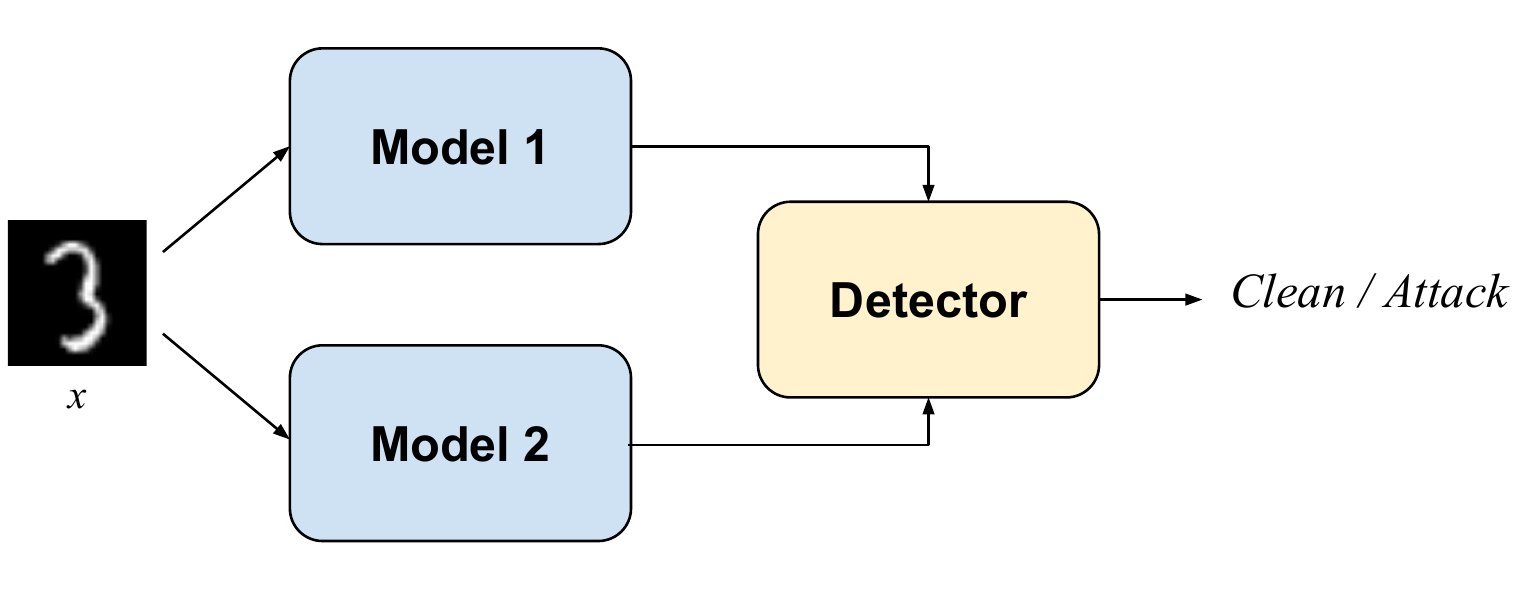}
\caption{Detection method using decision mismatch - concatenated outputs of pretrained Models 1 and 2 are used as features for the detector.}
\label{fig:detect}
\end{figure}

In this paper, we focus on detection, as being able to determine whether a system is under attack is a desired property in several applications, such as user authentication, access control, among many others. We exploit the fact that two different neural networks presented with the same attacker will make different mistakes and we propose to map such mistake patterns into a final detection, as depicted by Fig.~\ref{fig:detect}. Empirical evidence is provided supporting the claim that such strategy is effective in detecting several recently-introduced attack strategies as well as quality degradation attacks on two image classification tasks. More importantly, generalization to unseen attack strategies is assessed and the proposed detection method is shown to be able to attain accuracy higher than 90.0\% even when attack generation strategies never seen by the detector are used at test time.

The remainder of this paper is organized as follows: the proposed approach is motivated and detailed in Section~\ref{sec:proposedmethodforAEDetection}. Experimental setup is presented in Section~\ref{sec:experiments}, results and discussions in Section~\ref{sec:results}, and conclusions in Section~\ref{sec:conclusion}.

\section{Decision Mismatch for Attacks Detection: Motivation and Method}
\label{sec:proposedmethodforAEDetection}

It is known that different models make different mistakes when presented with the same attack inputs, and in order to exploit that, Feinman \textit{et al.} \cite{Feinman2017} use Dropout in ``train mode'' and sample distinct models to predict the same input. A divergence threshold is used to decide whether the input is an attack or not. However, models obtained by simply dropping out parameters of a given neural network share most of their structure, hence yielding similar gradient-based attacks. Using a similar strategy, in \cite{Xu2017} subsequent outputs of the same model obtained with an input sample before and after decreasing each pixel's color bit depth are used as features for binary classification of attacks vs. clean data. Once more, the same model is used for predictions of gradient-based attacks, which exploit the model structure for building attacks. Based on this, we argue that mismatching of predictions will be maximized when the model structure is relevantly distinct. We thus propose to perform classification with two independently trained models and use their softmax output as features for a separate binary classifier. Intuitively, the consistency of the outputs provided by the two systems will be an indicator of clean samples, while a mismatch will indicate potential attacks. Employing models with different architectures will make gradient-based attacks have different effects on each model, which is exactly what we intend to leverage in order to spot attacks.

The proposed approach is illustrated in Fig.~\ref{fig:detect}. Two independently trained multi-class classifiers (models 1 and 2) receive the same input, and their concatenated outputs are used as features by a binary classifier that performs classification of authentic vs. attack inputs. The requirements for model 1 and 2 are matching performance on the task at hand and significant structural differences. The detector can be any binary classifier of choice. The proposed approach also presents benefits in terms of cost since only two predictions are required at test time as opposed to e.g., \cite{Feinman2017}, which needed a relatively large number of predictions for divergence evaluation. Additionally, no extra processing is required on the tested samples, as was the case of \cite{Xu2017}. At train time our method requires the training of two different models performing the same task.

\section{Experimental Setup}
\label{sec:experiments}

\subsection{Model/detector training}

We perform experiments on the MNIST \cite{lecun1998mnist} and CIFAR10 \cite{krizhevsky2014cifar} datasets. MNIST contains 28x28 ten-class grayscale images representing the digits from $0$ to $9$. CIFAR10 is a ten-class dataset with 32x32 color images consisting of animals and vehicles. In both cases, 50,000/10,000 training/testing images are available. Two classifiers are trained in advance on each dataset (models 1 and 2 in Fig.~\ref{fig:detect}). Neural networks were implemented using Pytorch\footnote{\url{https://pytorch.org/}} \cite{pytorch}, while binary classifiers use Scikit learn \cite{pedregosa2011scikit}. The code to reproduce the experiments and results will be available on GitHub \footnote{\url{https://github.com/joaomonteirof}} upon acceptance. 
\\
\\
\textbf{Target models}: In the case of MNIST, models 1 and 2 were selected as convolutional (CNN) and fully connected (MLP) neural networks, respectively. Dropout regularization was employed in both models \cite{srivastava2014dropout}. The architectures are described in more details as follows: 

\begin{itemize}
\item \textit{CNN}: \textit{conv}(5x5, 10) $\rightarrow$ \textit{maxpool}(2x2) $\rightarrow$ \textit{conv}(5x5, 20) $\rightarrow$ \textit{maxpool}(2x2) $\rightarrow$ \textit{linear}(350, 50) $\rightarrow$ \textit{dropout}(0.5) $\rightarrow$ \textit{linear}(50, 10) $\rightarrow$ \textit{softmax}.
\item \textit{MLP}: \textit{linear}(784, 320) $\rightarrow$ \textit{dropout}(0.5) $\rightarrow$ \textit{linear}(320, 50)$ \rightarrow$ \textit{dropout}(0.5) $\rightarrow$ \textit{linear}(50, 10)$ \rightarrow$ \textit{softmax}
\end{itemize}
All activation functions were set to ReLU. Training was performed with Stochastic Gradient Descent (SGD) using a fixed learning rate of $0.01$ and mini-batches of size $64$. Training was executed for 10 epochs. 

In the case of CIFAR10, we selected the widely used convolutional neural networks VGG16 and ResNet50 \cite{simonyan2014very, he2016deep} as model 1 and 2 respectively. The models were trained from scratch using SGD. The learning rate was scheduled to start at $0.1$ and decay by a factor of $1/10$ after $10$ epochs with no improvement on the validation set accuracy. Momentum and L2 regularization were employed with respective coefficients set to $0.9$ and $0.0005$. Performance of the models for both datasets are presented in Table \ref{table:modelsperformance}. 

\begin{table}[h]
\centering
\caption{Performance on test data of the selected models.}
\label{table:modelsperformance}
\resizebox{0.9\columnwidth}{!}{
\tiny{
\begin{tabular}{ccc}
\hline
Dataset                  & Model & Test Accuracy (\%) \\ \hline
\multirow{2}{*}{\textbf{MNIST}}   & CNN            & 97.57                \\  
                                  & MLP            & 96.62                \\ \hline 
\multirow{2}{*}{\textbf{CIFAR10}} & VGG            & 92.50                \\  
                                  & ResNet         & 92.04                \\ \hline 
\end{tabular}
}}
\end{table}

\textbf{Training the attack detector}: In order to train the binary classifier (attack detector), subsets containing 10,000 images of the train partition of both MNIST and CIFAR10 were randomly sampled. We thus iterate over each such images and randomly decide whether it will be either included in the detector training data as a clean sample, in which case it receives a label of 0, or an attack will be generated on top of it. In this case, the sample will be included in the dataset with a label of 1. To generate attacks, one of the two models is randomly selected to be attacked, which means that all attacks considered are white-box for one of the targeted models and black-box with respect to the other. Moreover, we highlight that all attacks are generated with a variable budget in terms of how far from the original image an attack can be, as opposed to simply keeping a fixed budget across all examples, i.e. the distortion level is gradually increased up until the model makes a mistake, and this is independently repeated for each image. This is done so as to find effective attacks as close as possible to their original images, which will in turn be more difficult to detect. If the maximum distortion budget is used and a mistake is not made by the model, or if the model wrongly predicts a given image without any distortion, its clean version is included in the dataset with a label of 0. Four recently introduced gradient-based attack strategies were evaluated in this study, namely FGSM \cite{Goodfellow2015}, IGSM \cite{Kurakin2016}, JSMA \cite{Papernot2016}, and DeepFool \cite{Moosavi2016}. Moreover, quality degradation attacks were also considered, consisting of Gaussian Blur, Gaussian Noise, and Salt and Pepper. For reproducibility, all attacks were implemented using the open source tool Foolbox \cite{rauber2017foolbox}. 

\begin{figure}[h]
  \centering
  \includegraphics[width=0.48\textwidth]{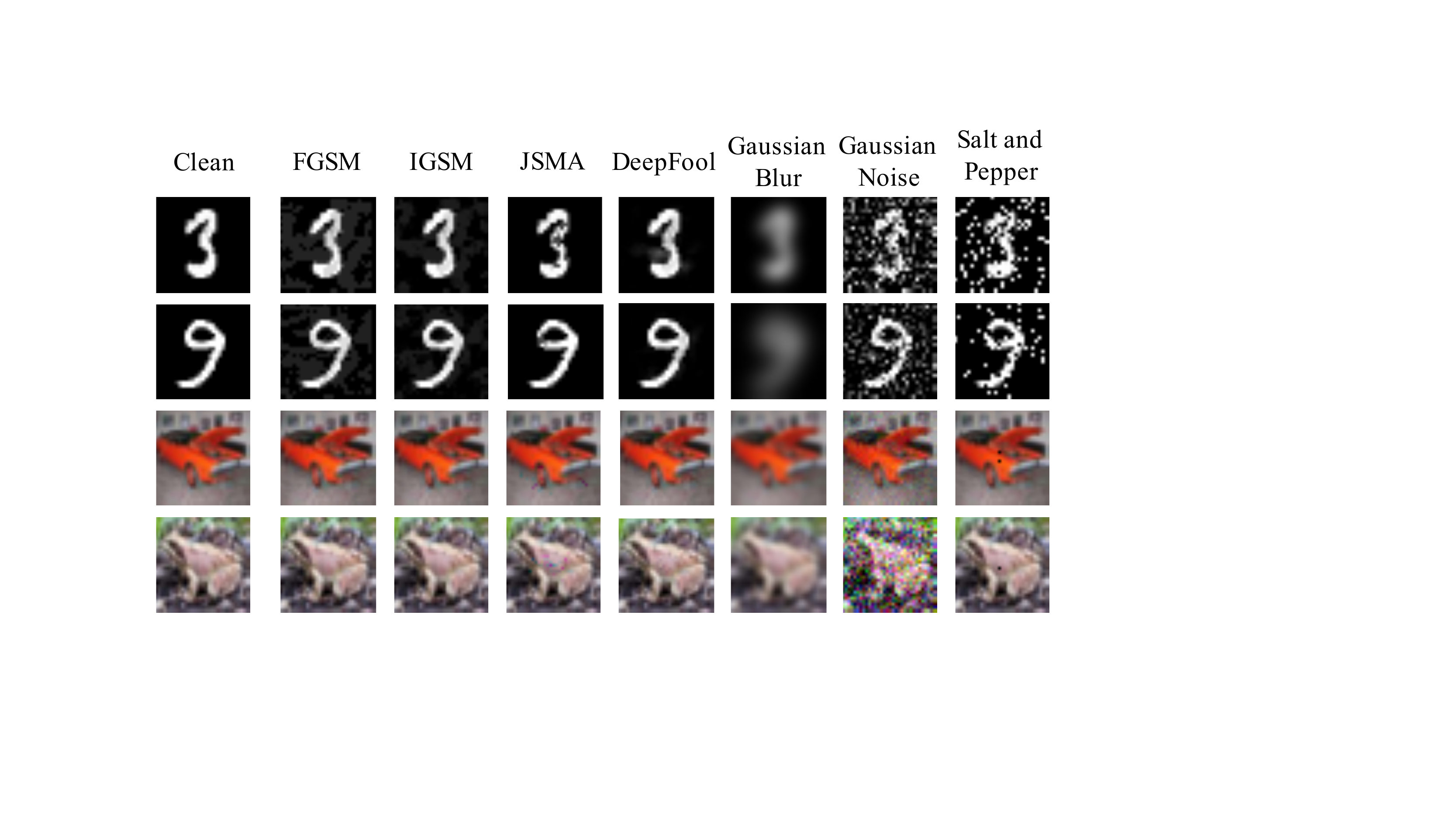}
  \caption{Examples of different attack strategies considered.}   
  \label{fig:samples}        
\end{figure}

In summary, by the end of the described process we obtained 7 independent datasets for each of MNIST and CIFAR10, each of those containing 10,000 samples being approximately 50\% attacks that were generated using either model 1 or 2. Such datasets are then used for training and evaluation of binary classifiers trained to classify data samples into genuine/attackers. Examples of above mentioned attacks are shown in Fig.~\ref{fig:samples}.

\subsection{Adversarial examples generation}
We now briefly describe the standard adversarial attack generation methods utilized in the experiments. Adversarial attacks to deep learning based systems can be either \emph{black-box} or \emph{white-box}, where the adversary, respectively, does not have and has knowledge of the model architecture, parameters and its training data. In addition, the attacks could be \emph{targeted} and \emph{non-targeted}, that aim to misguide deep neural networks to a specific class and arbitrary class except the correct one, respectively. In this work, we have used white-box non-targeted adversarial examples.

Let $f : \mathbb{R}^n \rightarrow \{1,...,K\}$ be an artificial neural network trained for multi-class classification, i.e., to map a given input $x \in \mathbb{R}$ to a class label $l \in \{1,...,K\}$ which might or might not correspond to $x$'s true class label $y$, where $K$ is the number of classes. Generating an adversarial attack can be thus defined as the minimization problem of finding an adversary $x^{adv}$ within a $p$-norm ball centered in $x$ which will yield an incorrect prediction, i.e. $f(x^{adv})=l^{adv} \not=y$, or, formally:

\begin{equation}
\begin{aligned}
\min_{x^{adv}} \quad &||{x^{adv}-x}||_p \\
\text{s.t.} \quad & f(x)= y \\
             & f(x^{adv})\not=y. \\
\end{aligned}
\label{eq:adv_gen}
\end{equation}

Several strategies were proposed to compute effective attackers. In the following, we provide a brief description of the attack strategies considered in our evaluation.
\\
\\
\textbf{Fast Gradient Sign Method \textnormal{(\textbf{FGSM})}}. FGSM \cite{Goodfellow2015} computes adversarial attacks by perturbing each pixel of a clean sample $x$ by $\epsilon$ on the direction of the gradient of the training loss $J$ (usually the categorical cross-entropy) given the true label $y$ with respect to the input:

\begin{equation}
 x^{adv} = x + \epsilon \cdot \textnormal{sign} [\nabla_x J (x,y)].   
\end{equation}

In this case, $\epsilon$ corresponds to the attack budget, i.e. the allowed distortion on $x$ to yield $x^{adv}$. A high $\epsilon$ will increase the attacker's success rate, but it will be easier to detect.
\\
\\
\textbf{Iterative Gradient Sign Method \textnormal{(\textbf{IGSM})}}. IGSM \cite{Kurakin2016} corresponds to iteratively repeating FGSM with a small $\epsilon$. The attack budget now depends on both $\epsilon$ and the number of iterations. $x^{adv}$ at iteration $i$ will be given by, for $x^{adv}_0 = x$:

\begin{equation}
    x^{adv}_i = x^{adv}_{i-1} + \epsilon \cdot \textnormal{sign}[\nabla_{x^{adv}_{i-1}} J (x^{adv}_{i-1},y)].
\end{equation}
\\
\textbf{Jacobian Saliency Map Attack \textnormal{(\textbf{JSMA})}}. In JSMA \cite{Papernot2016}, the goal is to determine which pixels yield higher variations on the outputs after perturbed. The outputs' Jacobian with respect to inputs is thus computed, and the perturbed value of each pixel $p$ of a given input $x$ belonging to class $y$, given the model $f$ and target class $t$, will then be $p^{adv} = p$ if
\begin{equation}
\frac{\partial f_t(x)}{\partial p} < 0 \hspace{1mm}\text{or} \hspace{1mm} \sum\limits_{j\not=t}\frac{\partial f_y(x)}{\partial p}>0.    
\end{equation}
Otherwise, 
\begin{equation}
p^{adv} = p + \frac{\partial f_t(x)}{\partial p} \vert \sum\limits_{j\not=t}\frac{\partial f_y(x)}{\partial p}\vert.    
\end{equation}
\\
\textbf{DeepFool}. For the binary case, i.e. K=2, DeepFool \cite{Moosavi2016} attacks are obtained by computing a linear approximation of the decision boundary, and then performing a step in its orthogonal direction, with the aim of finding the smallest perturbation yielding a change in the prediction given by the targeted model. The perturbation $r_i=x^{adv}_i-x$ at iteration $i$ will be given by:

\begin{equation}
    r_i = -\frac{f(x_i)}{||\nabla f(x_i)||^2_2} \nabla f(x_i),
\end{equation}
and $x^{adv}_{i+1}=x^{adv}_{i}+r_i$. Iterative updates of $x^{adv}_{i}$ stop when $f(x^{adv}_{i+1})\not=f(x^{adv}_{i})$. For $K>2$, a one-vs-all approach is employed and the direction corresponding to the minimum $||r_i||_2$ given each decision boundary is thus used for perturbing $x_i^{adv}$.
\\
\\
\textbf{Quality degradation attacks}. In addition to the aforementioned methods, we also produced adversarial attacks using quality perturbations, i.e., additive Gaussian noise, salt and pepper noise, and Gaussian blur were employed, and a line-search was performed to estimate minimal perturbations yielding a change in a given model's prediction.

\section{Results and Discussion}
\label{sec:results}
We performed two main experiments aimed at evaluating: (a) the detection performance of the proposed approach when training and testing are performed using the same attack strategy; and (b) the generalization capability of the proposed approach by using data generated with a particular attack strategy for training and a different one for testing the detector.

\subsection{Validation}
Prior to proceeding to main experiments results, we analyze the performance of the studied attacks and budget selection strategies. Ideally, powerful attacks will be close to their original images, and even so will be able to fool the target models. In Table~\ref{tab:attacks_perf}, the mean squared error (MSE) with respect to the original example is employed to measure the average distance between attacks and genuine images over random samples of size $500$ from both datasets. The success rate of each attack, i.e. which fraction of the $500$ images actually yielded a prediction mistake, is also reported.

In general, gradient-based methods yielded stronger attacks as the MSE of samples obtained with it is smaller for both MNIST and CIFAR. This fact is expected, as additive perturbation strategies methods do not directly exploit model structures to yield refined attacks. DeepFool is the strongest attack strategy evaluated on MNIST since its MSE is at least one order of magnitude lower than the other methods. For CIFAR10, IGSM is the best performer. FGSM is the weakest strategy among gradient-based attacks.

\begin{table}
\centering
\caption{Performance of different attacks evaluated in terms of the average MSE for 500 images and their success rate.}
\label{tab:attacks_perf}
\resizebox{0.485\textwidth}{!}{
\begin{tabular}{ccccc}
\hline
\multirow{2}{*}{Attack} & \multicolumn{2}{c}{MSE} & \multicolumn{2}{c}{Success Rate (\%)} \\
                        & MNIST      & CIFAR10    & MNIST           & CIFAR10        \\ \hline
FGSM                    & 2.13E-02   & 1.91E-03   & 100.0        & 100.0       \\
IGSM                    & 1.07E-02   & 3.34E-05   & 100.0        & 100.0       \\
JSMA                    & 1.24E-02   & 2.07E-04   & 95.6         & 100.0       \\
DeepFool                & 6.17E-03   & 1.00E-04   & 100.0        & 99.8        \\
Gaussian blur           & 4.35E-02   & 3.04E-03   & 90.2         & 93.8        \\
Gaussian noise          & 5.67E-02   & 4.90E-03   & 67.4         & 100.0       \\
Salt and pepper         & 1.23E-01   & 2.13E-03   & 100.0        & 100.0       \\ \hline
\end{tabular}
}
\end{table}

\begin{figure*}
	\centering
	\subfloat[tsne_raw][Raw images]{\includegraphics[width=0.5\textwidth]{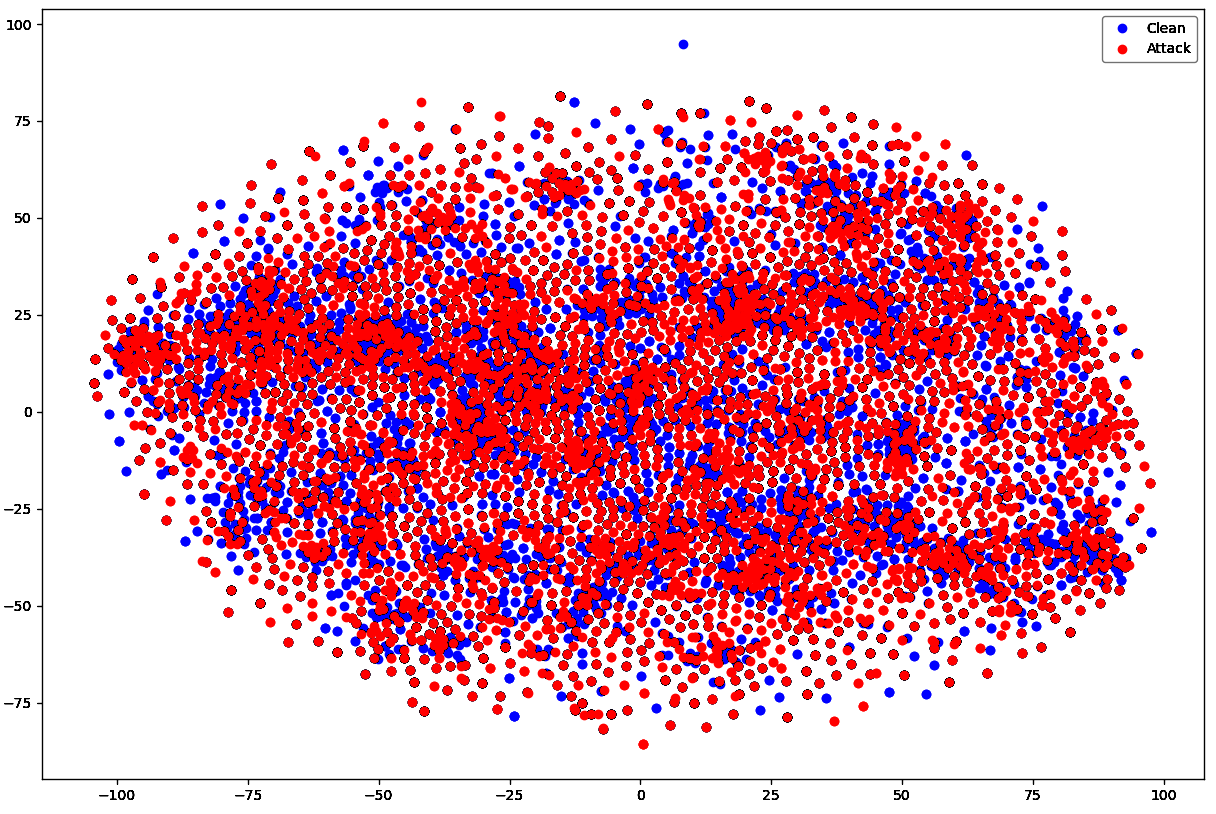}}
	\subfloat[tsne_ind][Features for independent models]{\includegraphics[width=0.5\textwidth]{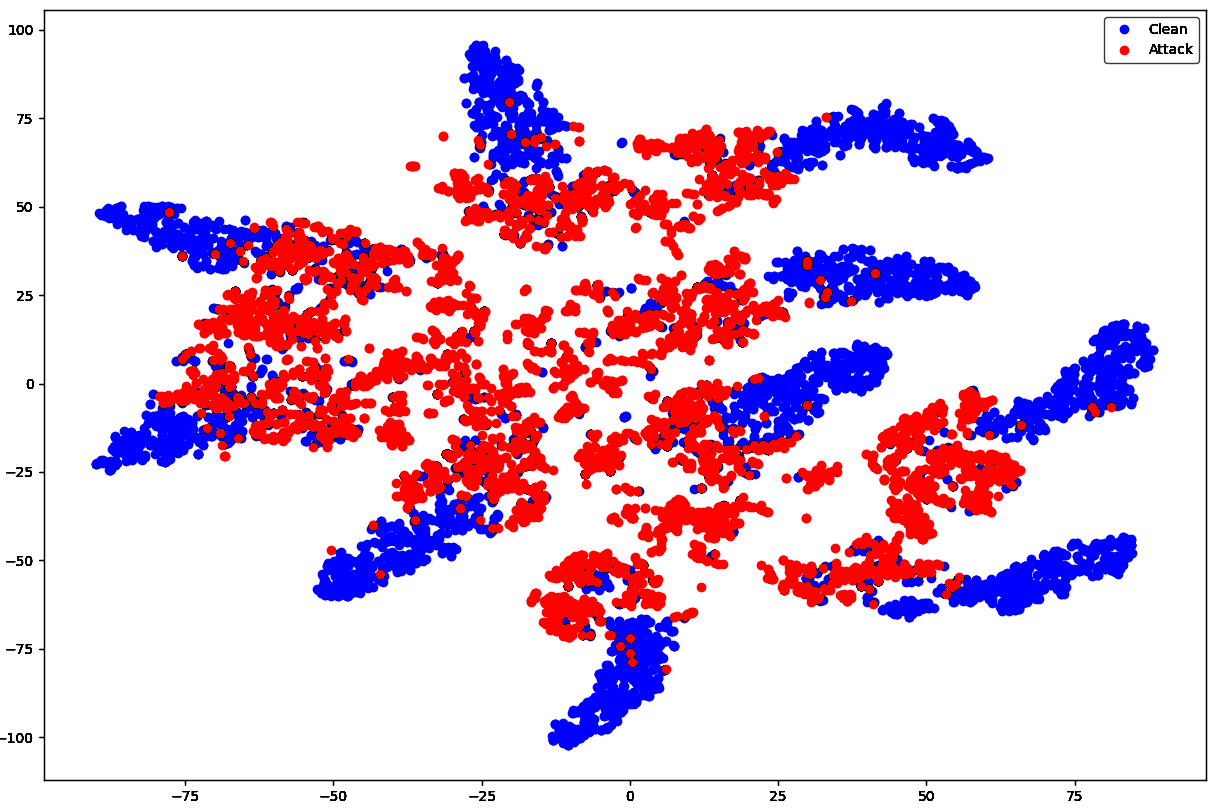}}'
	\caption{(a) Two dimensional t-SNE embeddings of raw images; (b) Features for independently trained model. Blue dots are authentic samples and red dots are attacks generated with DeepFool on 10,000 samples from CIFAR10. Attacks and clean data are indistinguishable for the case of raw images. Representations in the feature space make attacks and clean samples discriminable. Each blue cluster in the feature space corresponds to a particular class.}
	\label{fig:tsne}
\end{figure*}

Furthermore, to better visualize how discriminable attacks are when compared to clean samples, we show the low rank t-SNE embeddings \cite{maaten2008visualizing} of the raw images and their representations with our proposed features in Fig.~\ref{fig:tsne}. Plots are generated under DeepFool attacks on CIFAR10. One can clearly notice that the representation in the proposed space of features makes attacks and clean samples separable.

\begin{table}
\centering
\caption{SVM performance for (MNIST / CIFAR10) attacks detection using 10-fold cross-validation. Each experiment was performed with an independent random subset of the train data containing 10,000 images.}
\resizebox{0.49\textwidth}{!}{
\begin{tabular}{cccc}
\hline
                & Accuracy & F1     & AUC   \\ \hline
FGSM            & 0.93 / 0.93     & 0.92 / 0.93   & 0.93 / 0.93  \\
IGSM            & 0.91 / 0.93     & 0.91 / 0.93   & 0.91 / 0.93  \\
JSMA            & 0.92 / 0.93     & 0.92 / 0.92   & 0.92 / 0.93  \\
DeepFool        & 0.92 / 0.91     & 0.91 / 0.90   & 0.92 / 0.91  \\
Gaussian blur   & 0.98 / 0.97     & 0.98 / 0.97   & 0.98 / 0.97  \\
Gaussian noise  & 0.96 / 0.95     & 0.97 / 0.95   & 0.96 / 0.95  \\
Salt and pepper & 0.93 / 0.94     & 0.93 / 0.94   & 0.93 / 0.94  \\ \hline
\end{tabular}
}
\label{tab:mnist}
\end{table}


\begin{table}
\centering
\caption{Detection rate of attackers - Results are presented as (MNIST / CIFAR10). For MNIST, the proposed approach obtained detection rates on par with the Feature Squeezing (FS) and ensemble benchmarks.}
\label{tab:cifarcomp}
\resizebox{0.49\textwidth}{!}{

\begin{tabular}{cccc}
\hline
         & FS \cite{Xu2017}       & Ensemble \cite{bagnall2017} & Proposed                        \\ \hline
FGSM     & \textbf{1.000} / 0.208 & 0.998 / \textbf{0.998}      & 0.930 / 0.930                   \\
IGSM     & 0.979 / 0.550          & \textbf{0.997} / 0.488      & 0.920 / \textbf{0.940}          \\
JSMA      & \textbf{1.000} / 0.885 & -- / --                     & 0.930 / \textbf{0.970}          \\
DeepFool & -- / 0.774             & 0.450 / 0.426               & \textbf{0.910} / \textbf{0.910} \\ \hline
\end{tabular}}
\end{table}

\subsection{Attack detection performance}
To evaluate the proposed detection approach, we trained  a detector on top of the representations obtained with pretrained models using the previously described approach, i.e. concatenating the softmax layer outputs given by trained models. The binary classifier used here was a Support Vector Machine (SVM) with \textit{RBF} kernel. Accuracy, F1 score, accounting for precision and sensitivity simultaneously, and area under ROC curve (AUC) were analyzed as performance metrics. Evaluation of metrics was performed under 10-fold cross-validation, and results are shown in Table~\ref{tab:mnist} for MNIST and CIFAR10. As can be seen, scores are consistently higher than $90\%$.

In Table~\ref{tab:cifarcomp}, in turn, we provide an evaluation of our approach compared with results presented by Feature Squeezing (FS) \cite{Xu2017} and ensemble methods \cite{bagnall2017} in terms of detection rate or sensitivity, i.e., the fraction of attacks which were correctly classified by the binary classifier. Even though other methods present a higher sensitivity in the case of attack strategies such as FGSM and IGSM on MNIST, the proposed approach outperforms the benchmarks in more challenging cases, such as DeepFool or other attacks on CIFAR10, not showing any relevant performance degradation across methods. In the case of CIFAR10, our method outperformed the comparative techniques in 3 out of 4 attacks. We attribute the higher performance in those cases to the fact that the mismatch between the outputs of models 1 and 2 is higher for higher dimensional data and more effective attackers, and the detectors benefit from such higher mismatch.

\subsubsection{Generalization to unseen attacks}

\begin{figure}
	\centering
	\includegraphics[width=0.5\textwidth]{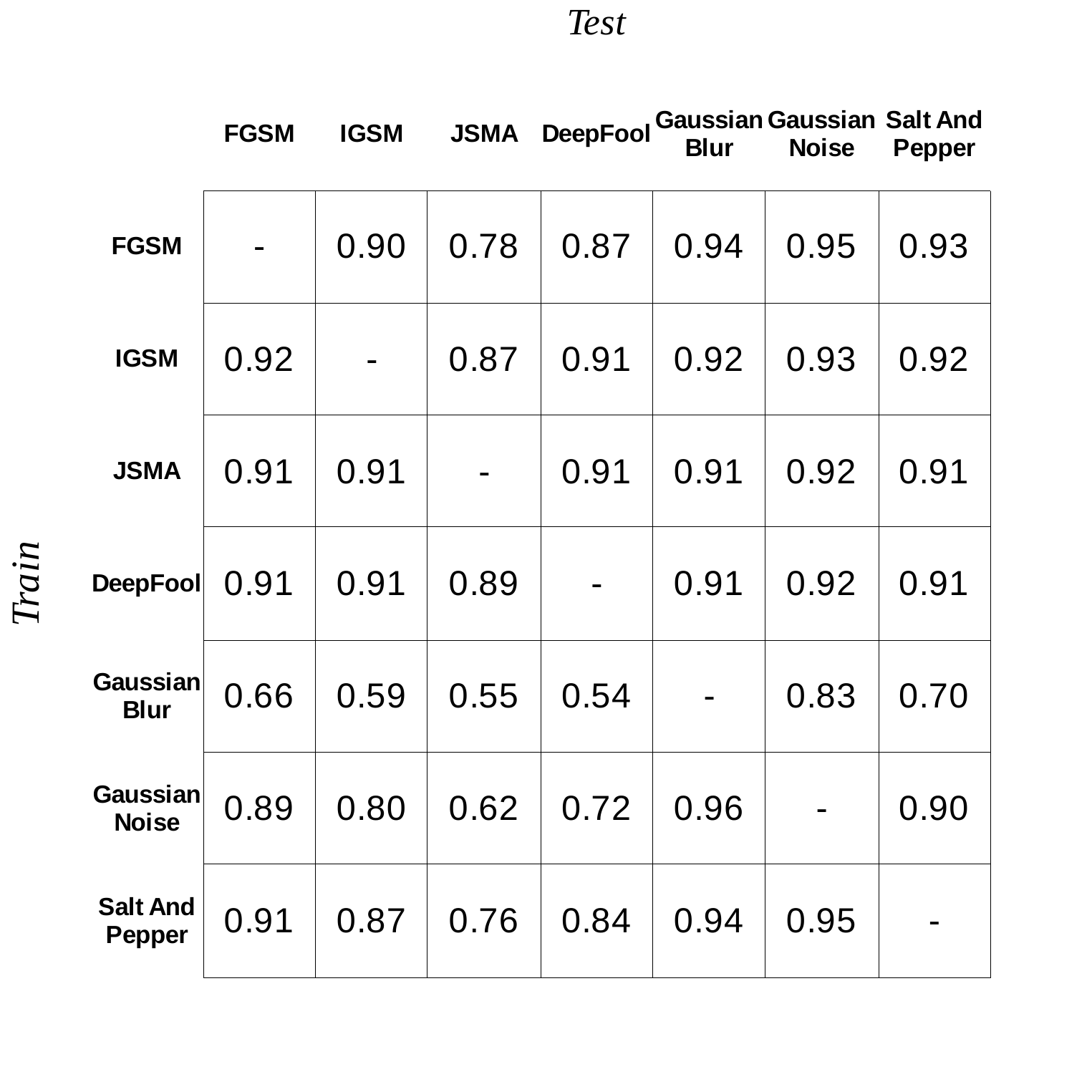}
	\caption{Generalization performance for MNIST of a SVM detector when trained in the attacks on the row and evaluated on the attacks on the columns. Values in the table indicate test accuracy - the higher the better.}
	\label{fig:mnist_gen}
\end{figure}

\begin{figure}
	\centering
	\includegraphics[width=0.5\textwidth]{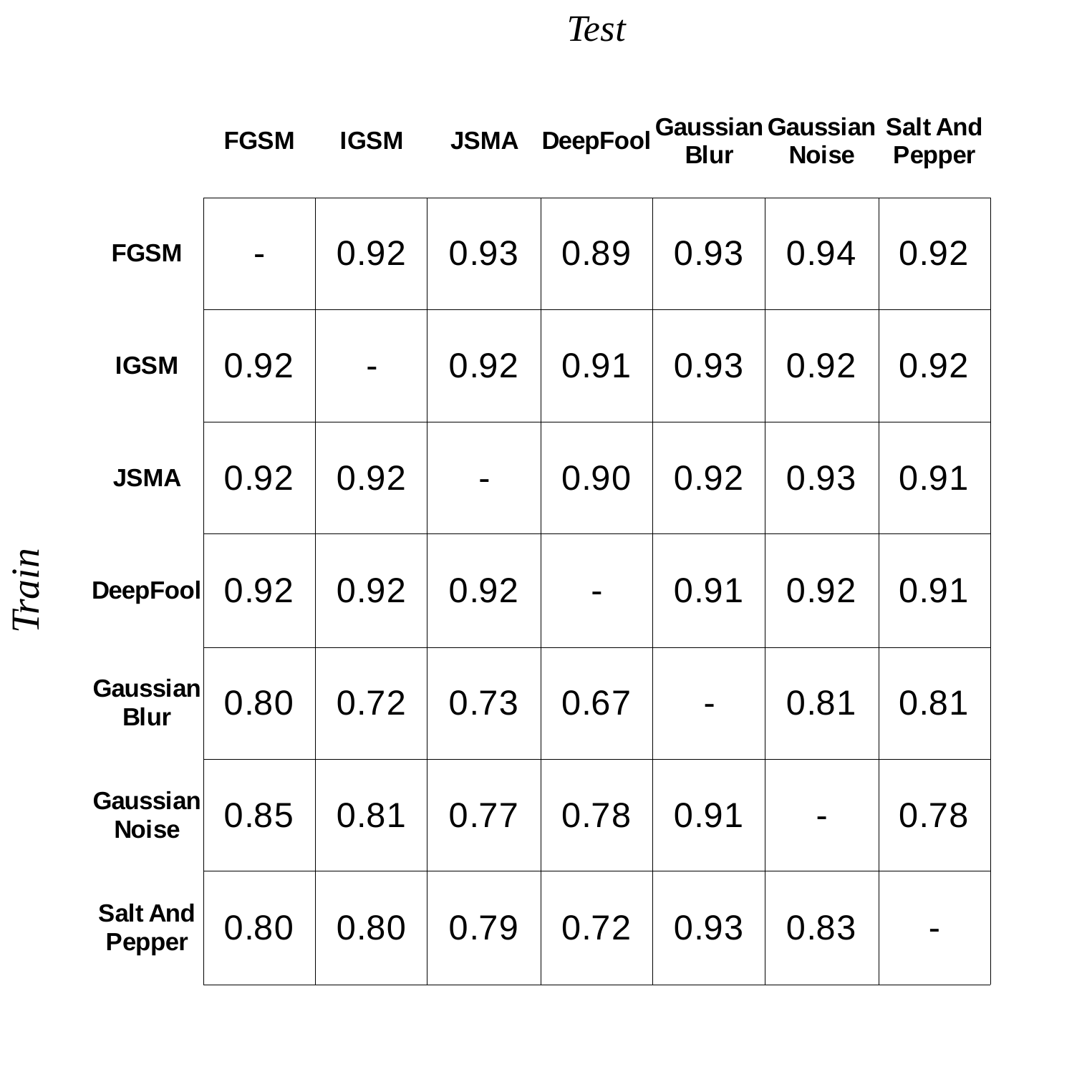}
	\caption{Generalization performance for CIFAR10 of a SVM detector when trained in the attacks on the row and evaluated on the attacks on the columns. Values in the table indicate test accuracy - the higher the better.}
	\label{fig:cifar_gen}
\end{figure}

The second experiment was performed with the goal of verifying whether a detector trained on a particular attack strategy is able to detect other types of attacks, which is interesting from  practical perspective, since in a real-world system, the nature of attackers is unknown, thus generalization across varying attack generation strategies arises as an important property of detection and defense schemes. In order to do so, we used data from a particular attacker for training the binary classifiers and evaluated their accuracy on different attack data. All possible pairs of attacks were evaluated under this scheme and the accuracies obtained are presented in Figures~\ref{fig:mnist_gen} and \ref{fig:cifar_gen} for MNIST and CIFAR10, respectively. Results indicate that the proposed approach presents high detection accuracy even when the detector is tested against unseen attacks. Detectors trained on JSMA on MNIST are the ones that perform best on different attacks. The same behavior appears on CIFAR10 for DeepFool. Interestingly, even though detectors trained on quality degradation attacks present a much lower generalization capacity, detectors trained on simple attacks such as salt and pepper noise can still achieve surprisingly high accuracies on refined attacks such as DeepFool and JSMA.

\section{Conclusion}
\label{sec:conclusion}

In this work we evaluated a novel approach to perform detection of adversarial attacks against artificial neural networks. To do so, the softmax layer outputs of two previously trained multi-class models are used as features for binary classification of attacks versus clean samples. Experiments presented provide empirical evidence that not only the space defined by the concatenated outputs of distinct pre-trained models is separable in terms of genuine examples and attacks, but more importantly that generalization to unseen attacking methods is achieved. Such findings are important, as new attack strategies can be conceived to outperform detection or defense methods that are only able to overcome attacks generated by strategies known in advance. One shortcoming of the proposed approach lies in the fact that it requires two models to be trained to perform the same task, which will incur cost overhead at test time. Nevertheless, this redundancy pays off in terms of added robustness, and can be also leveraged as an ensemble to improve performance on the task of interest when clean samples are given.

For future work, we intend to scale this approach to classification tasks involving a higher number of classes. Moreover, even though the unconstrained testbed employed here - i.e. untargeted attacks along with white-box model dependent distortions - yields the most subtle and hence difficult to detect adversarial examples, evaluating the proposed method on black-box and targeted attacks is a complementary analysis to be done.

\section*{Acknowledgment}

The authors wish to acknowledge funding from the Natural Sciences and Engineering Research Council of Canada (NSERC).

\bibliographystyle{IEEEtran}
\bibliography{bibliography}

\end{document}